\documentclass[12pt]{article}

\usepackage{times}

\usepackage{hyperref}
\usepackage{latexsym}
\usepackage{amssymb}
\usepackage{amsfonts}
\usepackage{amsmath}
\usepackage{subfig}
\usepackage{natbib}
\usepackage[utf8]{inputenc}
\usepackage{graphicx}
\usepackage{xcolor}
\definecolor{applegreen}{rgb}{0.0, 0.65, 0.31}
\definecolor{azure}{rgb}{0.0, 0.5, 1.0}
\usepackage{verbatim} % For using comment
\newcounter{lastnote}

\title{From text saliency to linguistic objects: \\
learning linguistic interpretable markers with a multi-channels convolutional architecture}

\author
{Laurent Vanni,$^{1,2}$ Marco Corneli,$^{2,3}$ Damon Mayaffre,$^{1}$ Frédéric Precioso$^{2,4}$\\
\\
\normalsize{$^{1}$Univ. C\^{o}te d'Azur,  BCL, UMR UNS-CNRS 7320, Nice, France}\\
\normalsize{$^{2}$ Inria, CNRS, Laboratoire J.A. Dieudonné, Maasai research team, Nice, France} \\
\normalsize{$^{3}$Univ. C\^{o}te d'Azur, Center of Modeling, Simulation and Interactions, Nice, France}\\
\normalsize{$^{4}$Univ. C\^{o}te d'Azur, I3S, UMR UNS-CNRS 7271, Nice, France}\\
}

\date{}

\begin{document} 


\baselineskip24pt


\maketitle

% In setting up this template for *Science* papers, we've used both
% the \section* command and the \paragraph* command for topical
% divisions.  Which you use will of course depend on the type of paper
% you're writing.  Review Articles tend to have displayed headings, for
% which \section* is more appropriate; Research Articles, when they have
% formal topical divisions at all, tend to signal them with bold text
% that runs into the paragraph, for which \paragraph* is the right
% choice.  Either way, use the asterisk (*) modifier, as shown, to

\begin{abstract}
A lot of effort is currently made to provide methods to analyze and understand deep neural network impressive performances for tasks such as image or text classification. These methods are mainly based on  visualizing the important input features taken into account by the network to build a decision.
However these techniques, let us cite LIME, SHAP, Grad-CAM, or TDS, require extra effort to interpret the visualization with respect to expert knowledge.
In this paper, we propose a novel approach to inspect  the hidden layers of a fitted CNN in order to extract interpretable linguistic objects from texts exploiting classification process. In particular, we detail a \emph{weighted} extension of the Text Deconvolution Saliency (wTDS) measure which can be used  to highlight the relevant features used by the CNN to perform the classification task.
%The intuitive formulation of the wTDS as well as its interpretability might 
%be of great interest for the linguistic community. 
We empirically demonstrate the efficiency of our approach on corpora from two different languages: English and French. On all datasets, wTDS automatically encodes complex linguistic objects based on co-occurrences and possibly on grammatical and syntax analysis.
%%%The relevant features detected by wTDS are coherent with those obtained by the popular explanation technique LIME. However the computational effort  required by wTDS is smaller since no sampling step is required.
\end{abstract}

\section{Introduction}\label{sec:intro}
Each author has a discursive identity made up of identifiable lexical and grammatical choices. Therefore, one of the challenges of deep learning on text is to describe these identities.

Although it was shown in the literature that, in terms of accuracy, CNN based approaches outperform existing classifiers based on statistical key-indicators (e.g. the relative words frequency) or other machine learning techniques, it is still not clear if and how CNNs make use of standard features used in text mining (for instance word co-occurrences). We might also go further and assume that, for text classification, CNNs can rely on other complex linguistic structures that might be of interest for linguists. In the attempt to shed some light on this topic, our approach mainly relies on deconvolution process (i.e. transpose process), allowing us to interpret the CNN features in the input space. 

This paper focuses on linguistic object analysis via a multichannel convolutional architecture. That is, a CNN is trained to associate several parts of transcribed political speeches to their speaker (e.g. E. Macron and D. Trump).  Our main contribution is an improvement of an existing measure, the Text Deconvolution Saliency (TDS)~\citep[TDS,][]{vanni2018textual}, called \emph{weighted} Text Deconvolution Saliency (wTDS), allowing us to visualize the linguistic markers used by the CNN to perform the classification of a text, but also to make them fully interpretable for the linguists. In order to have a relevant description of a dataset, the wTDS is included in a model that introduce two further contributions i) processing the CNN parameters in order to ``rank" text segments assigned to an author from the more to the less representative of that author and ii) introducing a multi-channel CNN architecture in order to exploit additional linguistic information (e.g. lemma or \emph{part-of-speech}) for each token.

The next section describes some of the most representative related works. Two of them are discussed in more details in order to motivate and better describe our own main contribution.

\subsection{Related works}\label{subsec:RW}
Since the seminal work of \citet{Coll:08}, adopting CNNs for several NLP tasks (part-of-speech tagging, chunking, named entity recognition and semantic labeling), many researchers have widely used CNNs for similar and other purposes, such as text modeling~\citep[e.g.][]{kalchbrenner2014convolutional} or sentence classification~\citep[e.g.][]{kim2014convolutional}.
While CNNs are not the only available deep architecture in Text Mining, it has been noticed that they have several advantages with respect to recurrent architectures (RNNs, in particular LSTM and GRU) when performing key-phrase recognition~\citep{yin2017comparative}.     
This supervised classification task is the one we are interested in this work. 
% Once more we underling that the recognition rates that can be obtained via CNNs, although impressive, are \emph{not} our main focus. Indeed,
In particular, we aim at uncovering linguistic patterns used to highlight \emph{similarities} and \emph{specificities}~\citep{Fel:07, Leb:98} in a corpus. Standard text analysis techniques originally relied on statistical scores, for instance on the relative frequency of words \citep[a.k.a. z-scores, see][]{lafon1980}. However, these techniques could not exploit more challenging linguistic features, such as syntactical motifs \citet{mellet09}. In order to overcome these limitations and to account for long term dependencies in sentences, CNNs have been recently used. Indeed, being CNNs more robust than RNNs to the vanishing gradient problem, they might be able to detect links between different parts of a sentence~\citep{dauphin2017language,wen2017network, adel2017global}. This property is crucial, since it was shown that long range dependencies emerge in real data~\citep{li2015visualizing}.
Aiming at inspecting these dependencies as long as other complex linguistic patterns, some tools explaining how CNNs perform the classification task are required.
In this regard, a recent crucial contribution is represented by the Local Interpretable Model-agnostic Explanations~\citep[LIME][]{lime2016} framework. The basic idea of LIME is to approximate any complex classifier (e.g. a CNN) by a simpler one (e.g. sparse linear) in a neighborhood of a training point $x_i$. 
A simplified representation $\tilde{x_i}$ of $x_i$ is adopted, and $N$  points in a neighborhood of $\tilde{x_i}$ are sampled uniformly and used to minimize a distance between the original classifier and the simpler one. Once the simpler classifier is trained, it can be used to assess the (positive or negative) contribution  of each feature to the classification task as easily as in linear models.
This approach provides very interesting results and is generic, since it can provide explanations for any kind classifier. However, for every training point it involves sampling $N$ neighbors and evaluating the classifier for each one of them. This might be computationally prohibitive, especially for high dimension data. 
In the context of key-phrase recognition, an alternative approach was proposed by~\citet{vanni2018textual}. 
They considered as input data text segments of fixed size ($M$ tokens).
Each data point was represented as an $M \times D$ matrix, where $D$ is the word embedding size. After training a CNN for an author recognition task, they used a Deconvolution Network~\citep{zeiler2014visualizing} to project the feature map back into the input data space. 
Thus, the ``deconvolution'' assigns to the $m$-th token in the $i$-th text segment (say $d_{im}$)  a vector $x_{im} \in \mathbb{R}^D$. The sum of its entries defines the Text Deconvolution Saliency (TDS) of $d_{im}$. Intuitively, the higher (respectively lower) the TDS of $d_{im}$, the more (less) $d_{im}$ contributed to assign the text segment to its class (i.e. its author).
Although this approach returns meaningful results it may suffer from some inconsistencies in the explanation, as it will be shown in Section~\ref{sec:CIL}. 
In order to preserve the computational efficiency of TDS (once the CNN is trained it can be computed at a cost of one model evaluation per data point) we propose an improved version of the TDS (Section~\ref{subsec:wTDS}) overcoming the explanation drawbacks. %Moreover, the CNN architecture used in this paper differs from the one detailed in \citet{vanni2018textual} due to i) the adoption of the multichannel lemmatization described in Section~\ref{subsec:DecovMultiLemm} and ii) the optimization of the Deconvolution Network 

%%%%%\textcolor{red}{In this paper, our contributions are three-fold:
%%%%%\begin{itemize}
%%%%%    \item 
%%%%%\end{itemize}}

This paper is organized as follows: Section~\ref{sec:CIL} describes our CNN architecture as well as our contributions. Section~\ref{sec:experiments} illustrates the framework described in Section~\ref{sec:CIL} on two datasets: a English corpus and a French corpus. Section~\ref{sec:conc} concludes the paper and outlines some perspectives for future research.

\section{Model and contributions}\label{sec:CIL}
The first part of this section details our model, a convolutional neural network, trained for author classification tasks. In this work, this task corresponds to an intermediate step but does \emph{not} represent our final goal. Indeed, the scope is to learn how to exploit a trained CNN to recover linguistic markers, specific to the different authors. 
Thus, after detailing the architecture, we focus on some original contributions to the linguistic features extraction.
Our main contribution,  the \emph{weighted Text Deconvolution Saliency} (wTDS) is described in Section~\ref{subsec:wTDS}.
Two other contributions,  the \emph{softmax breakdown ranking} and the \emph{multi-channel convolutional lemmatization} are discussed in Section~\ref{subsec:softmax_mcl}.
\paragraph*{Notation.} In the following, $v \in \mathbb{R}^N$  will denote a real vector $v$ with $N$ entries. If not differently stated, it is intended to be a column vector. 
The notation $A \in \mathbb{R}^{M \times N}$ will be used to define a real matrix with $M$ rows and $N$ columns and the function $relu(\cdot)$ is defined as
\begin{equation*}
relu(x) = \max\{0,x\}
\end{equation*}

\subsection{CNN baseline}\label{subsec:baseline}
\begin{figure}[h]
\begin{center}
\includegraphics[width=7.8cm]{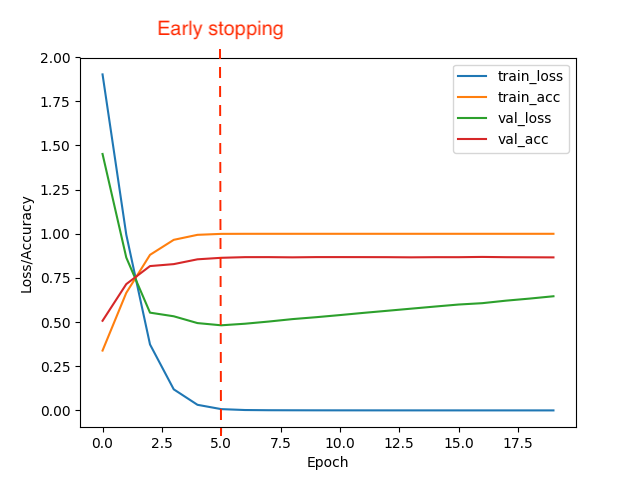}
\caption{Model loss and accuracy}
\label{fig:history}
\end{center}
\end{figure}

\begin{figure}[h]
\begin{center}
\includegraphics[width=7.8cm]{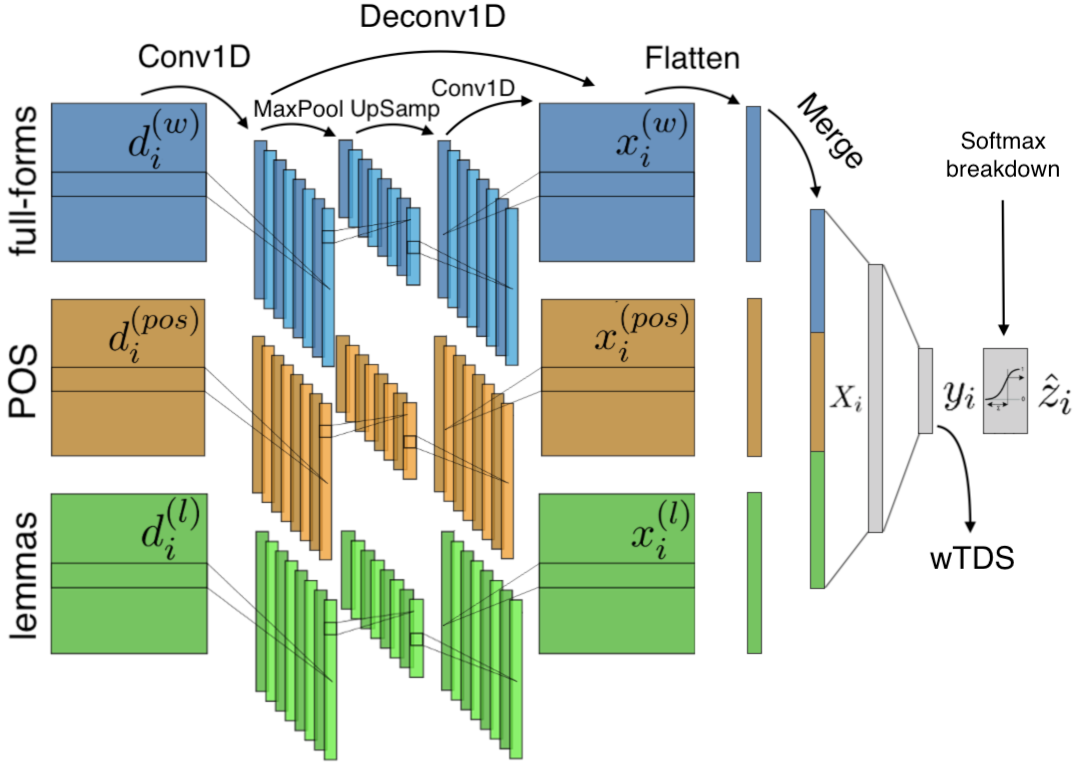}
\caption{Three channels convolution/deconvolution for three representation of the input 1) \emph{full-forms} (words), 2) \emph{part-of-speech} (POS), 3) \emph{lemma}}
\label{fig:MODEL}
\end{center}
\end{figure}

The CNN considered takes as input $d_1, \dots, d_N$ text segments, each containing a fixed number of tokens $M$.
In the examples that we consider in Section~\ref{sec:experiments} each segment is part of a presidential speech, so that the number of classes $K$ is the number of considered presidents.
An embedding layer is used for word representation.
Although this layer might rely on different well known models such as fastText~\citep{bojanowski2017enriching, joulin2017bag},  Word2Vec~\citep{mikolov2013distributed} or Glove~\citep{pennington2014glove} as long as a fine tuning of the embedding vectors is allowed during optimization, the choice of the embedding model is not crucial. Once the word feature vectors are obtained, they are concatenated (by row) in such a way to form a matrix with $M$ rows. This resulting matrix can then be input into a convolutional layer applying several filters all having the same width as the dimension of the embedding matrix. One max pooling layer follows, equipped with a non linear activation function.
A deconvolutional layer (up-sampling + convolution with transpose filters) is then introduced to bring the convolutional features back into the word embedding space. Finally, two fully-connected layers and a softmax function output for each segment $d_i$ a vector $\hat{z}_i \in \{0,1\}^K$, where $K$ is the number of classes/authors. The following multinomial cross-entropy loss function is considered:
\begin{equation}
\mathcal{L}(\theta) := - \sum_{i=1}^N \sum_{k=1}^{K} z_{ik}\log \left(\hat{z}_{ik}(\theta)\right)
\label{eq:loss}
\end{equation} 
where $\theta$ denotes the set of all the network trainable parameters and $z \in \mathbb{R}^{N \times K}$ is an observed binary matrix, whose $k$-th row encodes the class/author of the $i$-th text segment (thus $z_{ik} = 1$ iff $d_i$ is affected to the $k$-th class/author).
The above loss function is minimized with respect to $\theta$ via an Adam optimizer. 
In order to avoid overfitting the whole dataset is split into train ($80\%$) and validation ($20\%$) sets and the loss function in Eq.~\eqref{eq:loss} is monitored on the validation set during optimization, allowing us to apply early stopping~\citep{prechelt1998early} (Figure~\ref{fig:history}).
A graphical representation of the model described so far can be seen in Figure~\ref{fig:MODEL}. 

\subsection{A new enriched TDS}
\label{subsec:wTDS}
\begin{figure*}[t]
\begin{center}
\subfloat[TDS \label{fig:TDS}]{%
\includegraphics[width=.45\linewidth]{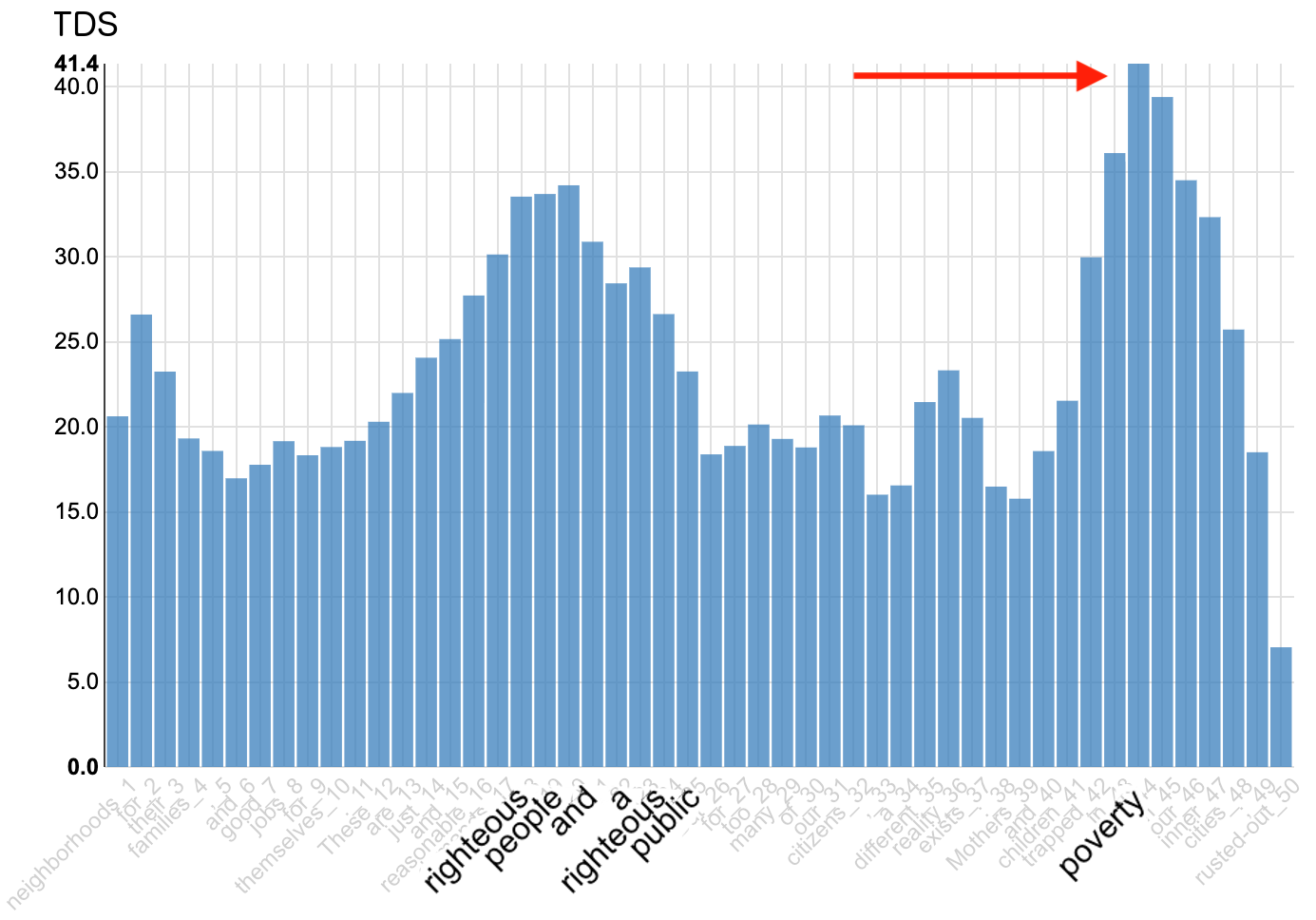}
}
\subfloat[LIME\label{fig:LIME}]{%
\includegraphics[width=.45\linewidth]{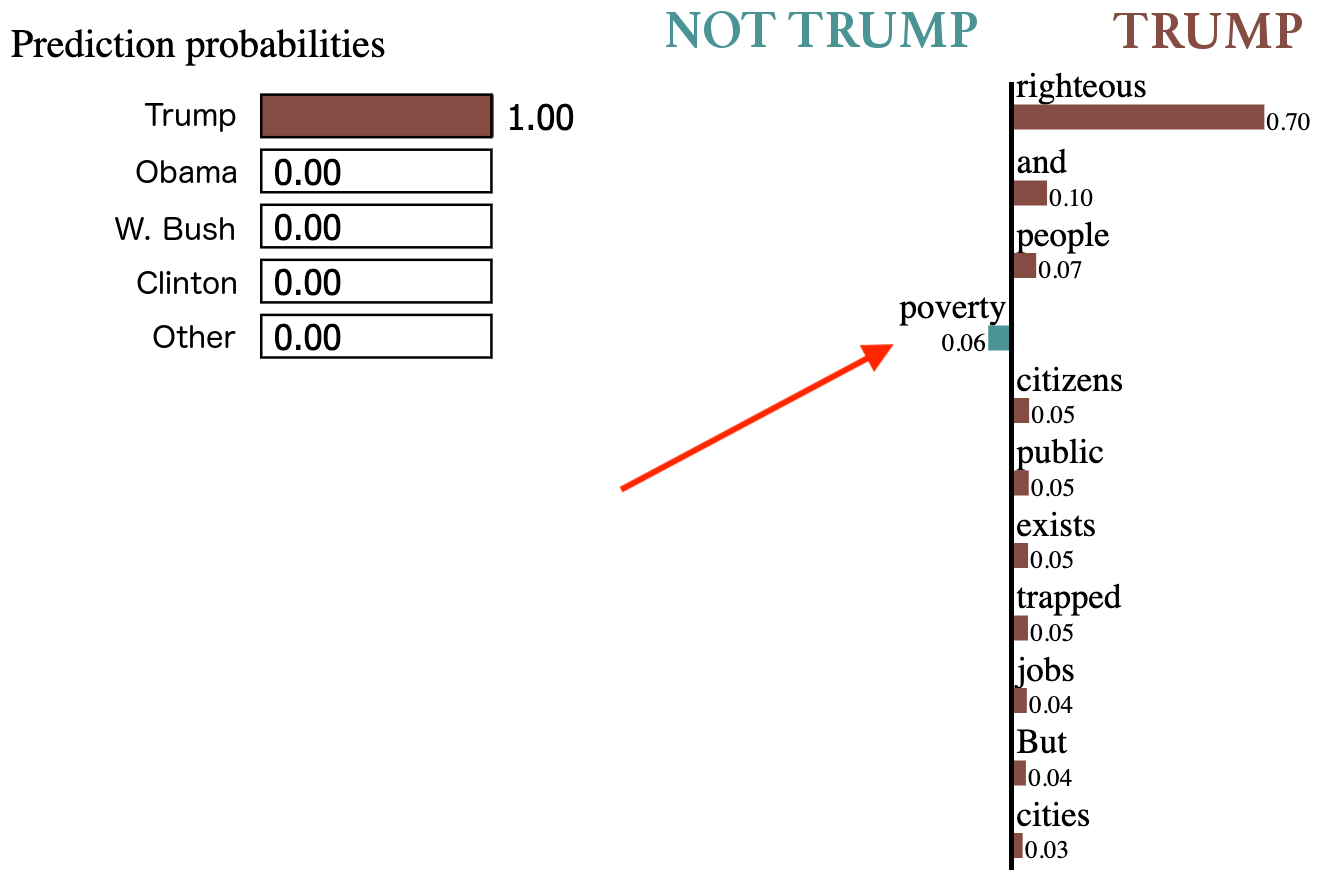}
}
\end{center}
\caption{Comparing the activation boost of the tokens toward the class ``Trump'' according to TDS and LIME.}
\end{figure*}
After the CNN has been trained on the train dataset, it can assign a text segment $d_i$ (either in the train or in the validation set) to its class/author. We recall that $d_i$ can be viewed as a real matrix with $M$ rows, where $M$ is the number of tokens of $d_i$ and $D$ columns, where $D$ is the embedding size. The $m$-th token of $d_i$, corresponding to the $m$-th row of the matrix, is denoted by $d_{im}$ and it is a vector in $\mathbb{R}^D$. The deconvolutional layer (see Figure~\ref{fig:MODEL}) assigns to every $d_{im}$ another vector of the same size denoted by $x_{im} \in \mathbb{R}^{D}$. Note that, since this representation is the output of two convolutional layers, it is sensitive to the context of $d_{im}$ (neighbor tokens). The Text Deconvolution Saliency~\citep[TDS,][]{vanni2018textual} of the token $d_{im}$ is defined as
\begin{equation}
TDS(d_{im}) = \sum_{d=1}^D x_{imd}
\label{eq:TDS}
\end{equation}
where the real number $x_{imd}$ is the $d$-th entry of $x_{im}$
%\footnote{Since the deconvolutional layer is equipped with a \emph{relu} activation function, $x_{imd} \geq 0$ for all $d$, thus we can adopt the following equivalent notation
%\begin{equation}
%TDS(d_{im}) = \rVert x_{im} \rVert_1
%\end{equation}
%where $\rVert \cdot \rVert_1$ denotes the $l_1$-norm of a vector.}.
We stress that, although this measure is defined for each token of $d_i$ it also accounts for the context of $d_i$ (see also the experiments in Section~\ref{sec:experiments}). 
The authors in~\citet{vanni2018textual} argue that, the higher the TDS of a token, the more the token (conditionally to its context) plays a crucial role in the classification task, according to the CNN. As  a matter of fact,
even though TDS can correctly highlight the relevant words/contexts in $d_i$ being used by the CNN to classify $d_i$, it cannot tell us \emph{how} the network uses them.
To illustrate this point in more detail, consider the following extract from a speech by Donald Trump:

\begin{quote}
\textit{[...] neighborhoods for their families , and good jobs for themselves . These are just and reasonable demands of  \textcolor{red}{righteous people and a righteous public} . But for too many of our citizens , a different reality exists : Mothers and children trapped in  \textcolor{red}{poverty} in our inner cities ; rusted-out [...]} 
\end{quote}

\noindent(D. Trump, the 20th of January 2017, Inaugural Address, United States Capitol Building in Washington, DC).

This text is part of a corpus described in Section~\ref{sec:experiments} and collects several part of speeches from the US presidents. Once properly trained for an author recognition task,
the CNN detailed in the previous section can correctly recognize this speech as being pronounced by the president Trump.
In Figure~\ref{fig:TDS} an histogram reports the TDS scores for the tokens of the extract.  The higher the bars, the more the corresponding tokens had a key role in the classification task.
Now, when comparing these TDSs with the word contributions detected by LIME (Figure~\ref{fig:LIME}) we see that most of the tokens having a high TDS correspond to  brown right bars having a \emph{positive} impact in classifying the speech as ``Trump" (e.g. righteous, people). Conversely, according to LIME, the noun ``poverty''  seems to have a negative boost when performing a binary classification ``Trump" or ``No Trump".
\begin{figure}[ht]
\begin{center}
\includegraphics[width=.9\linewidth]{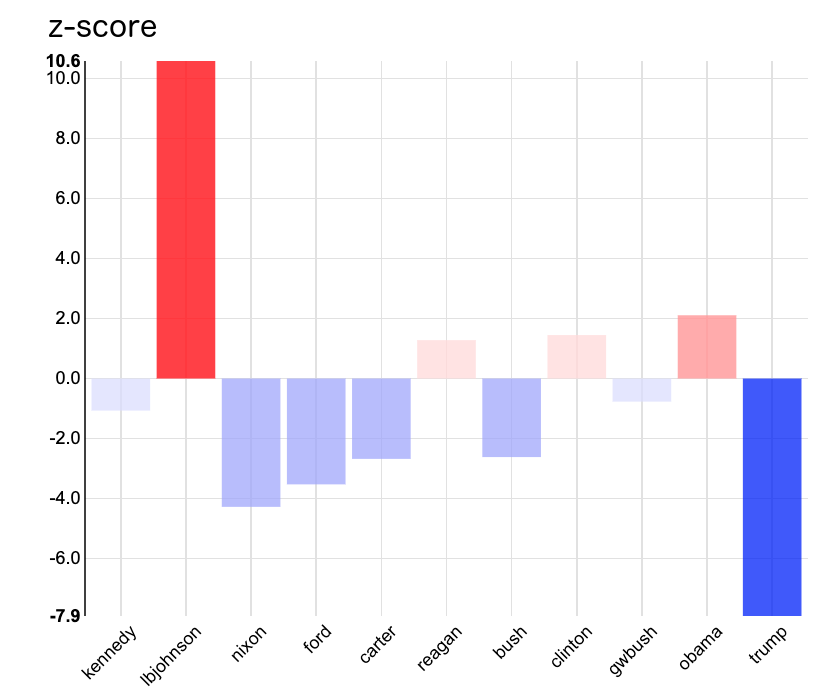}
\caption{z-scores for the noun ``poverty'' for the US presidents in the analyzed corpus.}
\label{fig:poverty}
\end{center}
\end{figure}
Indeed, if we additionally compute the z-scores of the tokens of $d_i$ (Figure~\ref{fig:poverty}), with respect to the whole corpus, we see that the noun ``poverty'' is underused by D. Trump and this is in line with the explanation provided by LIME. However, this noun is very specific to another president in the corpus: L.B Johnson.  Thus, the importance of the word ``poverty'' was correctly captured by TDS, but we cannot say if that word contributed \emph{for} ``Trump'' or \emph{against} ``Trump''.

\begin{figure*}[t]
\begin{center}
\subfloat[Trump\label{fig:Trump}]{
\includegraphics[width=.45\linewidth]{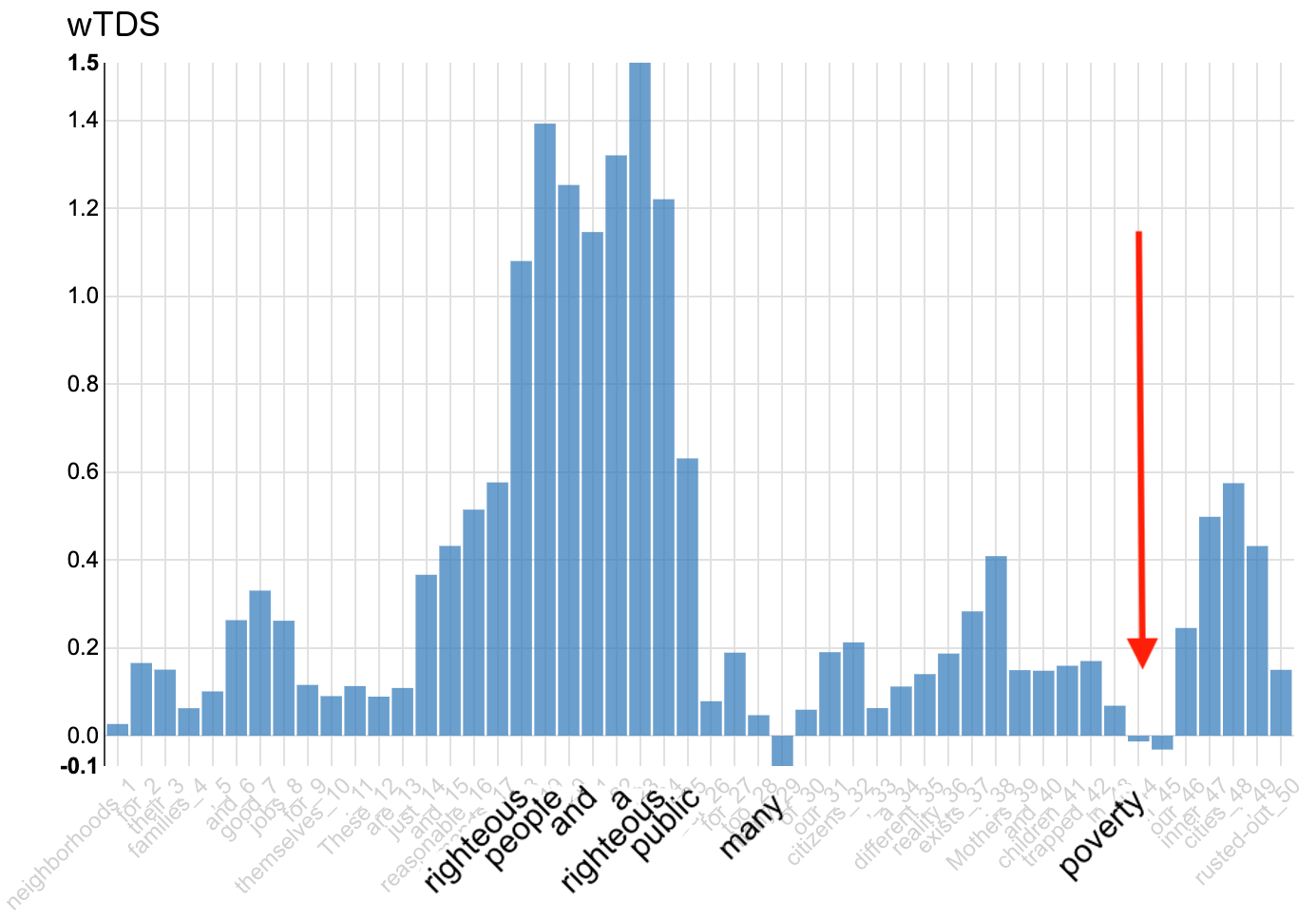}
}
\subfloat[Johnson\label{fig:Johnson}]{
\includegraphics[width=.45\linewidth]{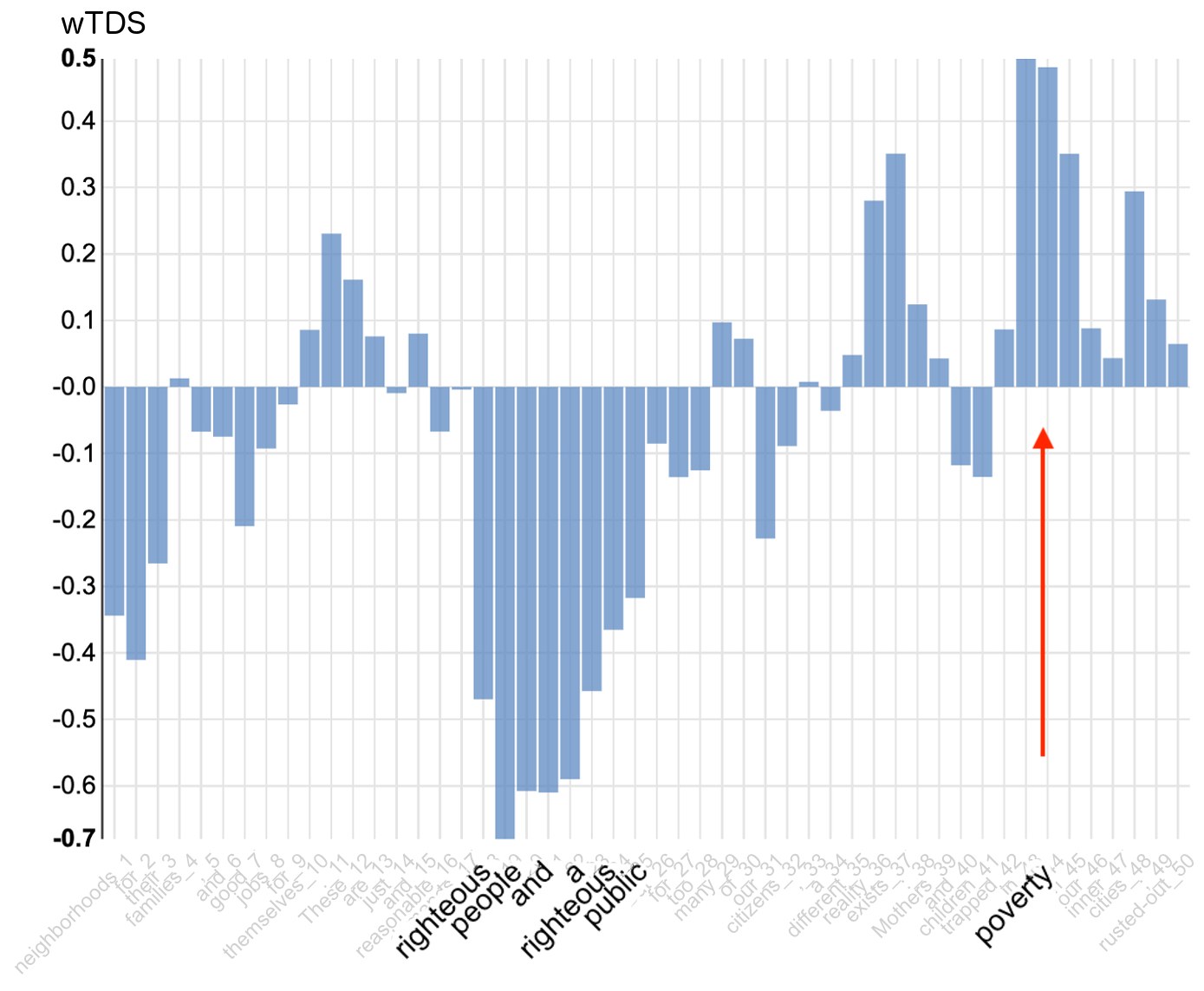}
}
\caption{wTDS for classes ``Trump'' and ``Johnson'' for the tokens in the sample speech of D. Trump.}
\label{fig:wTDS}
\end{center}
\end{figure*}

This motivated us to improve the TDS score initially proposed by~\citet{vanni2018textual}, with two additional features: i) it should be able to go negative to indicate negative contributions of words to some classes and ii) in case of multi-class classification, for a word $d_{im}$ it should be able to quantify its contribution to each class.
In order to build such a measure, note that the last two fully connected layers of the CNN basically map the de-convolved features $x_{i1}, \dots, x_{iM}$ into a single vector in $\mathbb{R}^K$, denoted $y_i$ (see Figure~\ref{fig:MODEL}), where $K$ is the number of classes.
If we concatenate $x_{i1}, \dots, x_{iM}$ into a column vector $X_i$, of size $D$x$M$, the map can be specified as
\begin{equation}
y_i = d + C\left( \text{relu}\left( b + AX_i \right) \right)
\label{eq:last_layer}
\end{equation}
where $A \in \mathbb{R}^{E \times DM}$, $b \in \mathbb{R}^{E}$, $C \in \mathbb{R}^{E \times K}$ and $d \in \mathbb{R}^{K}$ and $E$ is the size of the penultimate layer. In order to obtain a score that is specific to the token $d_{im}$ we observe that
\begin{equation}
AX_i = \sum_{m = 1}^M A_m x_{im}^T
\end{equation}
where $A_m \in \mathbb{R}^{E \times D}$ is the sub-matrix of $A$ obtained by selecting all the rows and the $D$ columns form the $(D(m-1) + 1)$-th to the $(D(m-1)+ D)$-th.
Thus we define
\begin{equation}
wTDS(d_{im}) := d + C\left(\text{relu}\left( b + A_m x_{im}^T \right)\right)
\label{eq:wDTS_def}
\end{equation}
Note that, instead of $TDS(d_{im})$, $wTDS(d_{im})$ is a vector with $K$ entries. Each entry quantifies the activation boost of word $d_{im}$ (conditionally to its context) for the class $K$. Moreover, the matrix multiplication $A_m x_{im}$ induced $K$ \emph{weighted} sums of the entries of $x_{im}$, in contrast with the simple sum defined in Eq.~\eqref{eq:TDS}. For this reason we call the measure in Eq.~\eqref{eq:wDTS_def} \textbf{weighted Text Deconvolution Saliency (wTDS)}. 
Figure~\ref{fig:wTDS} shows the wTDSs for the class ``Trump'' of the tokens in the Trump's speech reported above. 
As it can be seen, the word ``poverty'' now has a small negative contribution when classifying the speech as ``Trump''.
We notice that, once the CNN is trained, the computation of the wTDS for one token (for all the classes) has the cost of the matrix multiplications in Eq.~\eqref{eq:wDTS_def}. This is a huge advantage compared to LIME for two reasons: First, no sampling is required. Second, whereas LIME can only provide us with the tokens contribution in the binarized problem (e.g. ``Trump'' vs. ``No Trump'') , wTDS computes the tokens contribution to each class in one shot.

\subsection{Softmax breakdown ranking}\label{subsec:softmax_mcl}
In the previous section, we described how, given an input text segment $d_i$, wTDS can be used to assess the contribution of each token in $d_i$ for the class assignment.
Now, we zoom one step out and try to 
%Our main focus is to understand how the network accomplishes the classification task, in order to highlight the key-features characterizing the authors. 
%To do that, we first want to
 \emph{detect} the \emph{key-segments} in the data set, i.e. the segments being the more representative of each author according to the CNN. In particular, it might be of interest to be able to rank $d_1, \dots, d_N$ from the most to the least representative for each author.

A possible way to do that is described in the following. 
The number of neurons in the last layer of the deep CNN  coincides with the number of classes, previously denoted by $K$. In the previous section $y_i \in \mathbb{R}^K$ denoted the value of that layer for the  text segment $d_i$ . Thus, $y_{ik}$ is the value of the $k$-th neuron and it is a real number.
As usually, a softmax activation function is applied to $y_i$ in such a way to obtain $K$ probabilities $\hat{z}_{ik}$ (see Figure~\ref{fig:MODEL}) lying in the $K-1$ simplex 
\begin{equation}
\hat{z}_{ik} = \frac{\exp(y_{ik})}{\sum_{j=1}^K \exp(y_{ij})}
\end{equation}
Note that the above $\hat{z}_{ik}$ is the very same as in Eq.~\eqref{eq:loss}.
The highest probability $\hat{z}_{ik}$ corresponds to the class assigned by the network to the observation $d_i$.
However, if one entry of $y_i$ is significantly higher than the others, it is mapped to $1$ by the softmax transformation and all the other entries are mapped to zero.
For instance, consider two de-convolved features $y_i$ and $y_j$ corresponding to two different documents both assigned to class $k$. Assume also that $y_{ik}>y_{jk}$, so that the document $d_i$ is more representative of the class than $d_j$. If $y_{ik}$ and $y_{jk}$ are large enough, after applying the softmax function they both will be mapped to one and it will no longer be possible to assess whether $d_i$ or $d_j$ is more representative of class $k$.
Thus, we make unconventional use of the trained deep neural network and observe the activation rate of neurons \emph{before} applying the softmax transformation. Doing that, allows us to sort the learning data (text segments) based on their activation strengths. This simple but efficient method provides us with the most relevant \emph{key-segment} in the corpus for each class.
%We stress that, since the sorting described so fare is potentially performed on the entire data set $\{d_1, \dots, d_N\}$, overfitting is one of our main concerns. To avoid such an issue, we recall the whole corpus is split into train ($80\%$) and validation ($20\%$) data sets and the loss function in Eq.~\eqref{eq:loss} is monitored on the validation data set during optimization, allowing us to perform early stopping~\citep{prechelt1998early}. However, being  the training data set the largest part in our corpus, our analysis mostly rely on it.

\subsection{Multichannel convolutional lemmatization}
\label{subsec:Mcl}
Often, CNN  for images have multiple channels. Indeed, the RGB colors encoding could be considered as three different representations of the input. Each representation corresponds to a data matrix and the convolutional layers apply different filters to each matrix and then later merge the results. 
Also with texts, it is possible to encode the data in multiple channels that might be used, for instance, to combine different word embedding solutions (skip-gram, cBow or Glove). 
%It is also possible to fine-tune some channels in order to avoid the over-fitting~\citep{kim-2014}. 
Apart from word embedding, a pre-tagging process~\citep{Coll:08} allows data scientists and linguists to get supplementary material on each word, such as the \emph{part-of-speech} (POS) and the \emph{lemma}. Both of them are essential for a linguistic interpretation of the key-segments and to observe complex linguistic patterns \citep[a.k.a syntactical motifs][]{mellet09}. It is those reasons motivated us to implement a multi-channel CNN to account for the POS and the lemma.
However, using a single multi-channel convolutional layer to learn those patterns from each representation is not convenient for our purposes. Indeed, the max pooling operations merge all the information into one channel, thus making it impossible to retrieve which representation (word, POS or lemma) contributed to the classification. Since the aim of our contribution is to interpret the classifier, we split the convolution (and the max pooling) in three parts, one for each channel (see Figure~\ref{fig:MODEL}). 
By doing that, the deconvolution mechanism can be applied to the three channels separately and all the linguistic features can be observed right after the deconvolutional layers. Finally, to combine this information, the features are merged into a global vector and the final dense layers use them to perform the class assignment. 
In more details, the $m$-th token of the segment $d_{i}$ is now represented by three embedding vectors, say $d_{im}^{(w)}$ for the full form, $d_{im}^{(pos)}$ for the POS and $d_{im}^{(l)}$ for the lemma (see Figure~\ref{fig:MODEL}). After deconvolution, these embedding vectors are mapped to $x_{im}^{(w)}$, $x_{im}^{(pos)}$ and $x_{im}^{(l)}$, respectively. Thus, whereas with a single channel, $wTDS(d_{im})$ was a vector in $\mathbb{R}^K$, in a multichannel environment, we can define \textbf{three} wTDS vectors in $\mathbb{R}^K$ for each token. For instance, $wTDS(d_{im}^{(l)})$ refers to the lemma component of the $m$-th token and it can be computed as
\[
wTDS(d_{im}^{(l)}):= d + C\left(\text{relu}\left(b + A_m^{(l)} (x_{im}^{(l)})^T \right)\right)
\]
where $A_{m}^{(l)}$ denotes a sub-matrix accounting for the lemma channel (the green one in Figure~\ref{fig:MODEL}) \emph{and} the $m$-th token $x_{im}^{(l)}$.

\section{Experiments}
\label{sec:experiments}
First we want to thank the authors of TDS~\cite{vanni2018textual} for providing us with their datasets.

Political discourse analysis is one of the major challenges for linguistics in textual data analysis. For many years, statistics have provided tools and results that help linguists to interpret political speeches. We will now see how our deep architecture allows us to describe international political discourses. We propose to test our model by analysing two political discourse corpora in two different languages, English and French. For comparison reasons, these two corpora are made from presidential speeches and respect the same chronological span, from the 1960s to today.

The first dataset targets American political discourse. It is a corpus of 1.8 millions of words of American presidents from J.F. Kennedy in 1961 to D. Trump in 2019. With 11 presidents, we focus on D. Trump to make a short but profound  linguistic analysis of the discourse of the current US president. The second is symmetrical with the speeches of the French presidents under the 5th republic from 1958 to today. It is 8 French presidents from C. De Gaulle to E. Macron with 2.7 millions of words we focus also on current president, E. Macron.

By default, the accuracy of each model (English and French) exceeds 90\%, but the markers displayed by the wTDS seem to be too sensitive to low frequencies (very rare linguistic markers) or on the contrary very frequent but unique to a president (high z-score). The purpose of our architecture being to observe new linguistic markers different from those known by statistics, each corpus has been filtered with precise rules to reduce the weight of these markers. Some words have been replaced: i) proper names ii) dates iii) words only present in a president. These rules reduce model accuracy by about 10\% but help to reduce overfitting and extract relevant key segments. The table \ref{tab:corpus} compare those models, unfiltered (English, French) and filtered (English*, French*)

\begin{table}[h!]
\begin{center}
\begin{tabular}{|c|c|c|c|c|}
\hline
dataset & authors & vocab & words & acc  \\
\hline
English & 11 & 33279 & 1 815 839 &  90\% \\ 
English* & 11  &  14758 & 1 815 839 &  81\% \\ 
French & 8 & 46978 & 2 738 652 & 91\%  \\
French* & 8 & 20211 &  2 738 652 & 84\% \\
\hline
\end{tabular}
\caption{English and French datasets.}
\label{tab:corpus}
\end{center}
\end{table}

\subsection{English data set}\label{subsec:EnDatasets}

Section~\ref{subsec:wTDS}  introduce a key-segment of D. Trump detected with the \emph{softmax breakdown ranking} method with a simple model using only one channel for the full-form of words. With the \emph{multi-channel convolutional lemmatization} (Section~\ref{subsec:softmax_mcl}), we have now a wTDS score on each token for each channel and this selected segment become fully interpretable for the linguists due to exploitable features on full-form (blue words), \emph{part-of-speech} (orange words) and lemma (green words):

\begin{quote}
\textit{[...] neighborhoods for their families , and good jobs for themselves . These are just and reasonable demands \textcolor{applegreen}{of righteous people and a righteous} public \textcolor{orange}{SENT} \textcolor{azure}{But} for too many of our citizens , a different reality exists : Mothers and children trapped in poverty in our inner cities ; rusted-out [...]} 
\end{quote}

We highlight here the main activation zones having a wTDS higher than a fixed threshold. As it can be seen, there is a redundancy of ``righteous people'' and ``righteous public'', being part of a simple and compassionate vocabulary (e.g. ``families'', ``mothers'', ``children'' or simply ``good jobs''), which is typical of populist speeches.

``But'' appears as a characteristic of a polemical discourse that defines Trump's rhetoric. The president rarely makes a consensual speech. Opposition marks, as ``But'', allow him to build a speech setting him apart from the mainstream. Being ``But'' placed at the beginning of the sentence, its full-form wTDS highlights the role of conjunction of opposition rather than of conjunction of coordination.  

We also report that the full-form wTDS for the word ``many'' is negative (Figure \ref{fig:TDS}). Since ``many'' is one of the words more often employed by president Trump (high z-score), a negative wTDS might appear surprising. However in this context, ``many'' is preceded by ``too" which is taken into account by the convolution layer. Thus we checked the z-score of the linguistic pattern ``too many'', and we found out that it is higher for B. Obama than D. Trump. This is a very good example of the wTDS capability to capture the linguistic context.

\begin{figure}[h]
\begin{center}
\includegraphics[width=7.8cm, height=6cm]{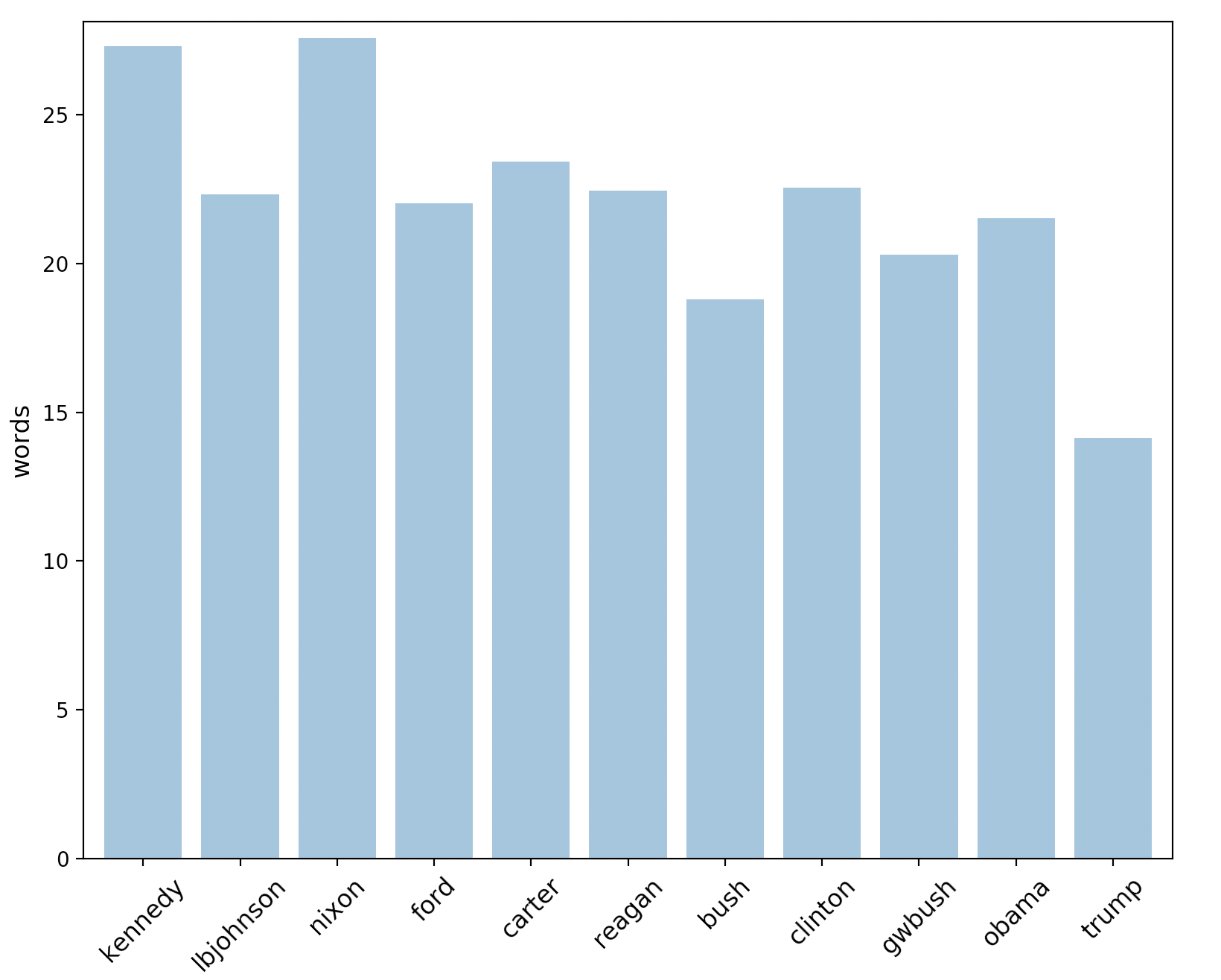}
\caption{Average sentence size}
\label{fig:sentence_size}
\end{center}
\end{figure}

Finally, the wTDSs of \emph{part-of-speech} focuses on a simple but essential marker, the dot (encoded as ``SENT''). The over use of this marker refers to a fundamental rhetorical choice of D. Trump: short sentences. 
The reduction of the sentence length is a trend that can be observed in most democracies in Europe or in USA. In the attempt to be accessible to as many people as possible, D. Trump's speech thus plays on syntactic simplification \citep{Norris2019}. For a long time, political discourse has imitated literature with long sentences and relative or subordinate proposals, but nowadays, political discourse imitates popular language with short sentences that include only one subject, one verb and one complement. On average, in the corpus, Trump's sentence counts 14.15 words when Obama's sentence counts 21.51 words (Figure~\ref{fig:sentence_size}). In fact, the end of sentence markers characterize the current president.

In 50 words here, Trump seems to take up the linguistic characteristics of populist discourse \citep{Oliver2016} as it is expressed in the United States and Europe at the beginning of the 21st century.

\subsection{French data set}\label{subsec:FrDatasets}
This section aims at demonstrating that Deep learning can easily adapt to the subtleties of each language.  A French presidential corpus is considered. In this dataset, the segment that the model identifies as being the most characteristic of E. Macron's speech gathers remarkable features of the current French president language.  The wTDSs highlight linguistic markers with multiple interpretations: 

\begin{quote}
\textit{ [...] intérêts industriels et qui \textcolor{applegreen}{construire le opacité} \textcolor{orange}{PRP PRP:det} décisions collectives qu' attendent \textcolor{azure}{nos concitoyens} . La cinquième clé de \textcolor{azure}{notre} souveraineté passe par  \textcolor{applegreen}{le numérique . ce} défi est aussi celui d' une \textcolor{applegreen}{transformation} profonde de nos économies , de nos sociétés , de \textcolor{applegreen}{notre imaginaire même} . La [...]} 
\end{quote}

(Macron, the 26th of September 2017, speech about Europe at the Sorbonne).
Some main features of the E. Macron's speech emerge. First, the French president tries to give a non-ideological and pragmatic talk oriented towards action, movement and efficiency \citep{Colen2019}. Thus, the lemmas ``construire'' (to build) and ``transformation'' are very meaningful of such a discourse whose main scope is to be dynamic. The word ``numérique'' (digital) is often at the heart of the speech of a president who talks about changes and who wants to show his technical modernity. 
Then, from a grammatical and syntactic point of view, most of the time, the ``PRP PRP:det'' sequence (meaning preposition + contracted article, in French) introduces adverbial phrases. Thus, E. Macron avoids the main topics %principles %(subjects) 
but he is precise with the modalities of the action. In E. Macron's speech, both the subject and the object are less important than 
%  it is less the subject or object that is important than 
the way of the proposed reforms. 
Finally, from a lexical point of view, the CNN seems to focus on ``concitoyens'' (fellow citizens) which allows E. Macron to avoid the term ``compatriots'', considered too nationalist in the 21st century, in the context of the European integration. A high wTDS also corresponds to the ``nos'' and ``notre'' (``our'' and ``ours'') forms as well to the lemma ``notre''. Indeed, the construction of a political ``we'' appears as the main rhetorical objective of a discourse that aims at gathering the people behind its leader.

\section{Conclusion and perspectives}\label{sec:conc}
%%%%%\subsection{Conclusion}
We have introduced and tested a new method to extract relevant linguistic objects characterizing the different classes/authors in a multi-class classification context. The main focus of the present work are the hidden layers of a trained CNN. In particular we introduced a measure (wTDS) which, entirely relying on the learned parameters, allowed us to detect the key words that, conditionally to their context, were used by the CNN to assign a text segment to its author. We have proposed a routine to rank the text segments from the most to the least representative for each author providing a new and different view in the author discourse analysis.
The way we propose to compute all these features internally to the network leads to a highly reduced computation cost (compared to LIME for instance) and thus allows us to design a multi-channel architecture accounting for \emph{part-of-speech} and the lemma leading to extract enriched linguistic objects at almost no cost.

%%%%%%\subsection{Perspectives}

The linguistic objects that we learn in this multi-class classification framework are those better discriminating one author \emph{with respect to} the others. In order to extract not only discriminative spatial linguistic objects (using CNNs) but to take into account the sequential generation of the discourse based on these linguistic objects, recurrent networks have to be considered. Some tools already explore the hidden layers of such architectures (e.g. LSTMVis\footnote{\url{http://lstm.seas.harvard.edu/}.}) and future works might focus on the combination of both approaches, for instance, first extracting spatial patterns then analyzing their sequential organization for an even more in depth discourse analysis.

\newpage

\bibliography{anthology,anthology2}
\bibliographystyle{apalike}

\appendix

%\section{Appendices}
%\label{sec:appendix}

%\section{Supplemental Material}
%\label{sec:supplemental}

\end{document}